\documentclass{elsarticle}
\usepackage{titlesec}
\usepackage{graphicx}
\usepackage{amsthm}
\usepackage{amssymb}
\usepackage{mathtools}
\usepackage{booktabs} 
\usepackage{array}
\usepackage{multirow} 
\usepackage{enumerate}
\usepackage{placeins}
\usepackage{varwidth}
\usepackage{soul}
\usepackage{undertilde}
\usepackage{algorithm}
\usepackage{algpseudocode}

\makeatletter
\def\ps@pprintTitle{%
	\let\@oddhead\@empty
	\let\@evenhead\@empty
	\let\@oddfoot\@empty
	\let\@evenfoot\@oddfoot
}
\makeatother
\graphicspath{{Images/}}
\begin{document}
\begin{abstract}
Selection of proper stocks, before allocating investment ratios, is always a crucial task for the investors. Presence 
of many influencing factors in stock performance have motivated researchers to adopt various Artificial Intelligence 
(AI) techniques to make this challenging task easier. In this paper a novel fuzzy expert system model is proposed to 
evaluate and rank the stocks under Bombay Stock Exchange (BSE). Dempster-Shafer (DS) evidence theory is used for 
the first time to automatically generate the consequents of the fuzzy rule base to reduce the effort in knowledge base 
development of the expert system. Later a portfolio optimization model is constructed where the objective function is 
considered as the ratio of the difference of fuzzy portfolio return and the risk free return to the weighted mean 
semi-variance of the assets that has been used. The model is solved by applying Ant Colony Optimization (ACO) algorithm 
by giving preference to the top ranked stocks. The performance of the model proved to be satisfactory for short-term 
investment period when compared with the recent performance of the stocks.            
 
\end{abstract}

\begin{frontmatter}
\title{Fuzzy Expert System for Stock Portfolio Selection: An Application to Bombay Stock Exchange}

\author[gsmt]{Gour Sundar Mitra Thakur}
\ead{cse.gsmt@gmail.com}
\address[gsmt]{Department of Artificial Intelligence and Machine Learning, Dr. B. C. Roy Engg. College, Durgapur, West Bengal, India}

\author[rb]{Rupak Bhattacharyya}
\ead{mathsrup@gmail.com}
\address[rb]{Department of Mathematics,Bijoy Krishna Girls' College, Howrah, West Bengal, India}

\author[sm]{Seema Sarkar (Mondal)}
\ead{seemasarkarmondal@yahoo.co.in}
\address[sm]{Department of Mathematics, National Institute of Technology Durgapur, West Bengal, India}
\end{frontmatter}

\let\Oldsection\section
\renewcommand{\section}{\FloatBarrier\Oldsection}

\let\Oldsubsection\subsection
\renewcommand{\subsection}{\FloatBarrier\Oldsubsection}
\let\Oldsubsubsection\subsubsection
\renewcommand{\subsubsection}{\FloatBarrier\Oldsubsubsection}
\section{Introduction}
Stock portfolio selection is basically an optimal allocation of investor' assets among certain number of stocks to 
provide maximum return with minimum risk. There are different types of factors which directly or indirectly influence 
the performance of stocks. This has made the task of stock selection and portfolio construction very challenging and 
uncertain for the stock market researchers since the very beginning. According to \cite{Markowitz1952} portfolio 
selection may include two stages: firstly, performances of different securities is observed with beliefs about their 
future performances and secondly, a proper choice of portfolio is made with relevant beliefs about future performances. 
The main aim in Modern Portfolio Theory (MPT) of investment is to maximize the expected return of the portfolio for a 
given amount of portfolio risk or  equivalently  minimize risk for a given level of expected return, by carefully 
choosing the proportions of various assets. In a ground-breaking work \cite{Markowitz1952}, Markowitz has quantified 
return as the mean and risk as the variance of the portfolio of securities. The relationship between risk and 
mean-variance are later examined in various researches like \cite{Zhou2000,Fang2006,Ogryczak2000,Huang2008}. The twin 
objectives of investors', profit maximization and risk minimization, are thus quantified. Though this theory has been 
widely accepted and adopted, it is being criticized since last few years. The main reason behind this is in MPT 
the efficiency of market is considered to be the basic assumption and obtaining information about markets in frequent 
intervals is costly and time consuming \cite{Grossman1980}. Computational burden caused by the quadratic 
utility functions and covariance matrix when number of stock increases, is another major concern in MPT. Another 
limitation of MPT is, it does not give importance to real investors' preferences \cite{Xidonas2009}. In practice, it 
is also found that investors prefer portfolios that lie behind the efficient frontier of Markowitz model even though 
they are dominated by other portfolios with respect to return and risk. From the above discussion it is obvious that 
some additional criteria should be added to the classical risk-return framework.

Thus portfolio selection problem is proved to be a multi-dimensional problem and to solve this inherent multi-criteria nature of this problem Multi-criteria Decision Making (MCDM) approach has been adopted in various researches like \cite{ Xidonas2011, Edwards2007, Abdollahzadeh2002,Siskos1989}. In spite of all these efforts researchers were facing difficulties in bringing efficiency in portfolios specially in dynamic uncertain environment. As a result the use of different AI techniques in stock selection and portfolio construction became inevitable due to their well known capability of handling uncertain and vague data. In literature we can find wide use of different AI and Soft Computing techniques to analyse and predict the performance of stocks in various stock exchanges around the world. In some researches \cite{Adebiyi2012ANN, Fernandez2007ANN, Ko2008ANN, Olatunji2011ANN, De2011ANN} efficient learning capability of Artificial Neural Network (ANN) is adopted to select stocks and construct portfolios whereas in some other researches \cite{Chen2009GAa, Chang2009GA, Chen2009GAb, Jiao2007GA,Chen2010GA} optimization capability of Genetic Algorithm is used for portfolio optimization. But the application of Fuzzy Logic and Fuzzy Set theory has become the most popular among other soft computing techniques, due to its uncertainty handling capability and the efficiency to incorporate vagueness in investors' preferences in portfolio construction \cite{Bermudez2007fuzzy, Bilbao2006fuzzy, Fasanghari2010fuzzy, Huang2008fuzzy, Tiryaki2005fuzzy}. 

Portfolio selection process can be divided into two stages \cite{Xidonas2009}. In the first stage, some stocks are 
selected to construct the portfolio and in the second stage, the percentage of the total value for each stock is 
identified. Use of fuzzy expert systems in the selection of stocks and equity have recently become very effective and 
popular. In a work Fasanghari et al. \cite{Fasanghari2010fuzzy} developed an fuzzy expert system to select superior 
stocks in Tehran Stock Exchange (TSE) by identifying $7$ factors influencing the stock market. A rule base of total 932 
rules was constructed for this purpose. Though no proper portfolio optimization model was used in this work, the 
outcome of the model proved to be satisfactory. But the major concern of this model can be the development time and 
cost due to repetitive expert interactions for the development of the model. Authors have used only fuzzy set theory to 
deal with the inherent uncertainty present in the rule base though fuzzy set theory is more effective in dealing with 
vagueness in the model than addressing inherent uncertainty of the model \cite{Rodriguez2012hesitant}. In another 
significant work Xidonas et al.\cite{Xidonas2009} developed an expert system for the selection of equity securities but 
the concerns remained more or less same as the previous one.

\begin{figure}[h]
\centering
\includegraphics[width=8cm,height=7cm]{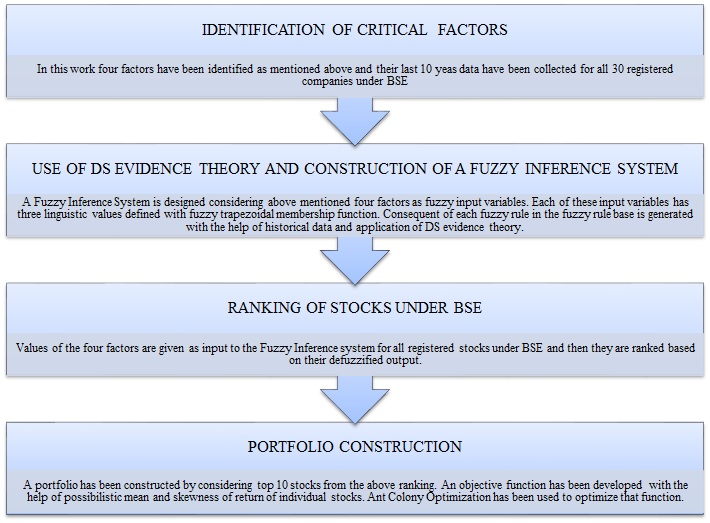}
\caption{Design diagram of the proposed inflation forecasting model}
\label{fig:F1}
\end{figure}

In this proposed work selection of stocks are made with the help of an expert system where the fuzzy rule base is 
automatically generated by applying DS evidence theory. This has considerably minimized the development time and cost 
as no cost is incurred here in terms of experts' consultations for the construction of the rule base. DS evidence  
theory is well known for its capability of dealing with incomplete and uncertain information. So another level of 
uncertainty handling mechanism beside fuzzy set theory is incorporated in the model. Though there are many 
factors which function as evidences of stock performance, here four most popular performance indicators: Price to 
Earning Ratio (P/E), Price to Book Value (P/B), Price to Sales ratio (P/S) and Long Term Debt to Equity ratio (LTDER) 
are initially identified and their historical data for all the registered stocks under BSE, are collected during the 
period of FY 2003-04 to FY 2011-12. In the first phase of the design, these four factors has functioned as input 
variables of 
a fuzzy inference system in the proposed model and their historical data have been used to generate the 
\emph{consequents} of the fuzzy rules with the help of DS evidence theory. The model is tested with the historical data 
of FY 2012-13, when values of these four factors for each stocks are provided as input to the fuzzy inference system 
and 
then they are ranked based on their \emph{defuzzified} output. In the second phase, a portfolio optimization model is 
constructed and then it is solved with ACO by considering top 10 stocks, as per the previous ranking, in the portfolio, 
to allocate investment ratios by giving preferences to the top ranked stocks. As mentioned earlier the model is tested 
with the data of FY 2012-13 to predict the performance of different stocks in FY 2013-14 and FY 2014-15, the 
performance of the model is proved to be satisfactory when compared with their actual performance. Figure \ref{fig:F1} 
depicts the brief structure of the article.

The rest of the paper is organized as follows: Section 2 gives brief description about the four identified factors and 
their importance in stock performance. In Section 3 design of the proposed fuzzy expert system model is elaborated. 
Portfolio construction using top 10 stocks is discussed in Section 4. Result analysis and comparison with recent 
performance of stocks is done in Section 5 and finally Section 6 concludes 
the discussion.

\section{	}  
There are many technical and fundamental factors which influence the stock market directly or indirectly. In this work four fundamental factors Price to Earning Ration (P/E), Price to Book Value (P/B), Price to Sales ratio (P/S) and Long Term Debt to Equity ratio (LTDER) are identified and their brief definitions are given below:

\subsection{Price to Earnings (P/E):}
The price to earnings ratio (P/E) is a valuation ratio of a stock that is determined by Market Value per Share divided by earnings per share (EPS) of the stock.
\begin{equation}
\frac{P}{E}=\frac{\textrm{Market Value per Share}}{EPS}
\end{equation}
In general, a high price to earnings ratio for any company indicates that investors are expecting higher earnings growth in future from it compared to companies with a lower price to earnings ratio.

\subsection{Price to Book value (P/B):}
Price to book value ration is used to compare market value of a stock to its Book value. It is also known as price to equity ratio. It is calculated by dividing the current closing price of the stock by the latest quarter's book value per share.
\begin{equation}
\frac{P}{B}=\frac{\textrm{Stock Price}}{\textrm{Total Assets}-\textrm{Intangible Assets and Liabilities}}
\end{equation}
In general, If P/B value is low then it suggests that the stock is undervalued and it could also mean that something is going wrong with the company.

\subsection{Price to Sales (P/S):}

Price to Sales ratio is another valuation ratio that compares a company's stock price to its revenues. It can be calculated by dividing the company's market capitalization by its total sale. This ratio is also known as revenue multiple or sales multiple.
\begin{equation}
\frac{P}{S}=\frac{\textrm{Market Capitalization}}{\textrm{Total Sales}}
\end{equation}
\subsection{Long term debt/equity (LTDER):}
Long term debt to equity ratio is the ratio of total liabilities of a stock to its shareholders' equity. It is a 
leverage ratio and it measures the degree to which the assets of the stock are financed by the debts and the 
shareholders' equity of a stock.  It is calculated by the total liabilities divided by shareholders' equity.

\begin{equation}
LTDER=\frac{\textrm{Total Liabilities}}{\textrm{Shareholders' Equity}}
\end{equation}

\section{DS-Fuzzy model for evaluation and ranking of stocks}
DS evidence theory and fuzzy rule based expert system are hybridized in this work to evaluate and rank the stocks under BSE. Brief introduction of DS evidence theory and fuzzy rule based expert system are given here before their application in the proposed model is elaborated. 

\subsection{DS evidence theory}
A. P. Dempster in 1967 \cite{Dempster1967} proposed a multi-valued mapping from one space to another which was used for statistical inference when we needed to identify a single hypothesis from multiple sample information. The DS-theory was first proposed by Dempster in 1968 \cite{Dempster1968} and later in 1976 it was extended by Shafer \cite{Shafer1976} to deal with incomplete and uncertain information.

DS-Theory mainly deals with four components: Frame of discernment, Basic Probability Assignment (BPA), plausibility function \emph{Pl} and belief function \emph{Bel}. 

In evidence theory first a set of hypothesises $\Theta$ called frame of discernment, is assumed and defined as follows:
\begin{equation}
\Theta=\{H_1,H_2,H_3,...,H_N\}.
\end{equation} 
The set is composed of N mutually exhaustive and exclusive hypothesises. From the $\Theta$, let us denote $P(\Theta)$, be the power set composed with $2^N$ propositions $X$of $\Theta$:
\begin{equation}
P(\Theta)=\{\varnothing, \{H_1\}, \{H_2\},..., \{H_N\},\{H_1\cup H_2\}, \{H_1\cup H_3\},...,\Theta \}
\end{equation}  
where $\varnothing$ denotes the empty set. The most important task in applying evidence theory is Basic Probability Assignment (BPA). A BPA is a function from $P(\Theta)$ to $[0, 1]$ defined by:

\begin{equation}
\begin{aligned}
m:&  &P(\Theta)\rightarrow [0,1]\\
  &  &X\mapsto m(A)
\end{aligned}
\end{equation}
and which satisfies the following criteria:
\begin{equation}
\begin{aligned}
\sum\limits_{X\in{P(\Theta)}}m(X)&=1;\\
m(\varnothing)&=1.
\end {aligned}
\end{equation}

Dempster's rule of combination is denoted by $m=m_1\oplus m_2$, which can combine two BPAs $m_1$ and $m_2$ to yield a new BPA:
\begin{equation}
\begin{aligned}
m(A)=\frac{\sum\limits_{Y\cap Z=X}m_1(Y)m_2(Z)}{1-k}
\end{aligned}
\end{equation}
with
\begin{equation}
\begin{aligned}
k=\sum\limits_{Y\cap Z=\varnothing}m_1(Y)m_2(Z)
\end{aligned}
\end{equation}
where k is a normalization constant, known as conflict because it measures the degree of conflict between $m_1$ and $m_2$. $k=0$ corresponds to the absence of conflict between $m_1$ and $m_2$, whereas $k=1$ implies total contradiction between them. The resulting belief function from the combination of $I$ information sources can be defined as:
\begin{equation}
m=m_1\oplus m_2 \oplus m_3 ... \oplus m_I
\end{equation} 
With the help of Equation (11) multi source information can be fused into a single framework in belief theory very 
easily and it proved to be very effective when it is applied in fuzzy rule generation discussed later in Section 3.3.3. 

\subsection{Fuzzy Expert System}
For many real world problems, like strategic planning, medical diagnosis and other decision making problems solutions are something more than simple reasoning and require some expertise knowledge to solve. So the use of Expert Systems (ES) became popular because it is capable of representing and reasoning knowledge rich domain with a view to solve problems and giving advice \cite{jackson1986introduction}. The main advantage of ES over any other conventional software applications is its effective reasoning capability as they process knowledge instead of data or information \cite{darlington2000essence}. On the other hand the application of fuzzy set theory in uncertainty handling has become one of the strongest tool to design any framework. As fuzzy set theory is now widely applied in information  gathering,  modelling, analysis, optimization, control, decision-making and supervision \cite{bellman1970decision}, fuzzy expert system is used in the construction of the proposed model.

\begin{figure}[h]
\centering
\includegraphics[width=8cm,height=5.5cm]{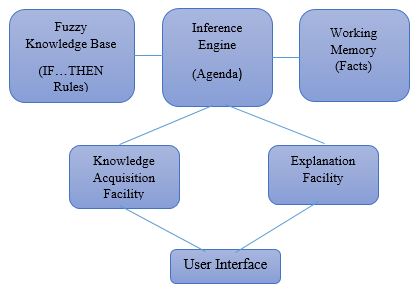}
\caption{Architecture of Fuzzy Rule-based Expert System}
\label{fig:F2}
\end{figure}

Fuzzy expert system is an expert system that uses fuzzy logic instead of Boolean logics in its knowledge base and derives conclusion from user inputs and fuzzy inference process \cite{kandel1991fuzzy}. Figure \ref{fig:F2} depicts the basic architecture of a fuzzy expert system.

Various components of a fuzzy expert system are briefly discussed below:
\begin{itemize}

\item \textbf{Fuzzy Knowledge Base}: Fuzzy knowledge base is a collection of fuzzy rules and it is built up by the knowledge engineer after eliciting knowledge from domain experts. The basic structure of a fuzzy rule is as follows:
\begin{equation*}
\text{IF } I_1 \text{ is } i_1 \text{ AND } I_2 \text{ is } i_2 \text{ AND ... AND } I_n \text{ is } i_n \text{ THEN } O \text{ is }o. 
\end{equation*}

where $I_1$, $I_2$ ... , $I_n$ are fuzzy linguistic input variables and $i_1$, $i_2$, ... , $i_n$ are possible linguistic values of $I_1$, $I_2$ ... , $I_n$ respectively. Similarly, $O$ is fuzzy linguistic output variable and $o$ is its linguistic value. The terms $I_1$ is $i_1$ and $I_2$ is $i_2$ ... are fuzzy propositions suggesting  partial  set  membership  or  partial  truth. In this way modifying the fuzzy sets parameters of the output fuzzy sets, the  partial  truth  of  the  rule premise can be evaluated \cite{matthews2003formal}.

\item \textbf{Fuzzy Inference Engine}: To  analyse  and manipulate  the  rules  in  the  knowledge  base an inference mechanism is adopted to conclude a logical result. Various inference mechanisms  can  be  adopted  for  fuzzy  inference  engine  depending on  the  aggregation, implications and operators used for s-norms and t-norms \cite{wang1999course}. Defuzzification of the results is also achieved  in this module. This modules functions as per the agenda. An agenda is a prioritized list of rules identified and enlisted by inference engine. Patterns of these rules are satisfied by facts or objects in the working memory. 

\item \textbf{Working Memory}: This module is contained of a global database of facts those are going to be used by the fuzzy rule base.

\item \textbf{Explanation Facility}: This particular module explains every steps in reasoning process of the ES.

\item \textbf{Knowledge Acquisition Facility}: This module provides an automated facility for the user to enter knowledge in the system.

\item \textbf{User Interface}: All kinds of interactions of the user with the system is facilitated by this module. This is used to accept inputs from the user and display the outcomes of the inference process.
\end{itemize}

\subsection{Design and Implementation of the proposed model}
The designed diagram of the proposed model is introduced in Section 1 as Figure \ref{fig:F1}. Here individual steps followed to construct and implement the whole model are elaborated.

\subsubsection{Collection of Historical Data}
 Total 30 stocks are registered under BSE. Initially, historical data of last 12 years (2003-04 to 2014-15) of four factors (P/E, P/B, P/S and LTDER) are collected from different web sources like www.capitaline.com, www.bseindia.com, www.nseindia.com etc. First 9 years' data are used for the implementation of the model and last 3 years' data are used for testing and comparison. For the simplicity of the model the data are normalized within the range of [0, 10] by scaling up/down the maximum historical value of any factor in last 9 years to 10 and then scaling other data accordingly relative to the maximum value. Table \ref{tab:Normalized_sample} shows an example of actual data and transferred for Axis Bank Ltd.
 
 \begin{table}[h]
 	\small
 \caption{Sample normalized data for Axis Bank Ltd.}
 \begin{center}
 \label{tab:Normalized_sample}
 \begin{tabular}{ | >{\centering\arraybackslash}m{4.2cm} | >{\centering\arraybackslash}m{3.25cm} | >{\centering\arraybackslash}m{3.25cm} | } 
 \hline
 \bf Financial Year & \bf Actual P/E ratio of Axis Bank Ltd. & \bf Transferred Data \\ 
 \hline 
 2011-12	& 11.46	& 4.24 \\
 \hline
 2010-11 &	17.5 &	6.48\\
 \hline
 2009-10 &	19.47 &	7.2\\
 \hline
 2008-09 &	8.49 &	3.14\\
 \hline
 \bf 2007-08 &	\bf 27.02 &	\bf 10\\
 \hline
 2006-07 &	21.67 &	8.02\\
 \hline
 2005-06 &	21.12 &	7.82\\
 \hline
 2004-05 &	20.5 &	7.59\\
 \hline
 2003-04 &	12.54 &	4.64\\
 \hline
 \end{tabular}
\end{center}
\end{table}

\subsubsection{Fuzzification}
Fuzzification is a process of taking a crisp value as input and transforming it into the degree required by the terms. Uncertainty basically comes from imprecision, vagueness or ambiguity and one of the best ways to represent these in any variable can be in the form of fuzzy variable. Fuzzy membership functions are used to represent the fuzzy variables. All four factors have been illustrated in TABLE II in terms of fuzzy linguistic variables and each of them has three linguistic values, \emph{Low, Standard and High}. Trapezoidal membership function has been used to represent the inputs within the range of [0, 10].

\begin{table}[h]
	\small
\caption{Membership functions for the linguistic values of the Input Variables}
\begin{center}
\label{tab:MFs}
\begin{tabular}{ | >{\centering\arraybackslash}m{4.05cm} | >{\centering\arraybackslash}m{3cm} | >{\centering\arraybackslash}m{3.5cm} |}
\hline
\bf Factors in terms of linguistic variables & \bf Linguistic Values & \bf Fuzzy Trapezoidal Membership\\
\hline
\multirow {3}{3cm}{\centering P/E Ratio (Range 0 to 10)} & Low & (0 \hspace{0.12cm} 0 \hspace{0.12cm} 1.8 \hspace{0.12cm} 2.8) \\ \cline{2-3}
& Standard &(1.7 \hspace{0.12cm} 3.5 \hspace{0.12cm} 4.6 \hspace{0.12cm} 5.8)\\ \cline{2-3}
& High & (5.3 \hspace{0.12cm} 7.5 \hspace{0.12cm} 10 \hspace{0.12cm} 10)\\ \cline{2-3}
\hline
\multirow {3}{3cm}{\centering P/B Ratio (Range 0 to 10)} & Low & (0 \hspace{0.12cm} 0 \hspace{0.12cm} 2.2 \hspace{0.12cm} 3.5)\\ \cline{2-3}
& Standard &(2.5 \hspace{0.12cm} 4.6 \hspace{0.12cm} 5.6 \hspace{0.12cm} 7.9)\\ \cline{2-3}
& High & (6.4 \hspace{0.12cm} 9.6 \hspace{0.12cm} 10 \hspace{0.12cm} 10)\\ \cline{2-3}
\hline
\multirow {3}{3cm}{\centering P/S Ratio (Range 0 to 10)} & Low & (0 \hspace{0.12cm} 0 \hspace{0.12cm} 2.4 \hspace{0.12cm} 3.6)\\ \cline{2-3}
& Standard &(1.8 \hspace{0.12cm} 4.4 \hspace{0.12cm} 5.7 \hspace{0.12cm} 8.2)\\ \cline{2-3}
& High & (6.4 \hspace{0.12cm} 8.7 \hspace{0.12cm} 10 \hspace{0.12cm} 10)\\ \cline{2-3}
\hline

\multirow {3}{3cm}{\centering LTDER (Range 0 to 10)} & Low & (0 \hspace{0.12cm} 0 \hspace{0.12cm} 2.3 \hspace{0.12cm} 3.6)\\ \cline{2-3}
& Standard &(2.8 \hspace{0.12cm} 4.6 \hspace{0.12cm} 5.7 \hspace{0.12cm} 7.6)\\ \cline{2-3}
& High & (6.124 \hspace{0.12cm} 8.09 \hspace{0.12cm} 10 \hspace{0.12cm} 10)\\ \cline{2-3}
\hline

\end{tabular}
\end{center}
\end{table}

\subsubsection{Fuzzy Rule Construction using DS-Theory}
The most significant novelty of this research work is the application of DS evidence theory in fuzzy rule construction. The knowledge base of any fuzzy inference system is developed using collection of fuzzy rules which determines how the output will be generated based on the given input. The proposed system has four input parameters namely, P/E, P/B, P/S and LTDER, and based on these input parameter the DS-Fuzzy system is expected to determine whether the selection of any stock will be \emph{Highly Favourable, Moderately Favourable or Not Favourable}, as output. The frame of discernment is considered as $\Theta=\{\text {\emph{High Performance}}, \text{\emph { Average Performance}}, \text{\emph { and Poor Performance}}\}$. 

Risk/Return ratio is one of the most commonly used measure to evaluate the performance of stocks. Risk is generally measured by semi-variance(S) of the stocks' previous returns and Return(R) is actually the mathematical mean of previous returns. Obviously lower value of this ratio indicate better performance of stocks. Based on linguistic values of each input factors and comparing those with S/R values as a measure of the performance of various stocks BPAs for different hypothesises are achieved. The historical data of FY 2012-13 are used to identify the stocks. Consulting various domain experts and analysing historical data initially standard values for each factors for all the stocks are decided. By analysing last 9 years' historical data it is found that when P/E ratio was around its standard value then $75\%$ of stocks under BSE performed better and S/R value was also found to be satisfactory. So a belief of 0.75 is assigned towards the hypothesis $\{\text{\emph{High Performance}}\}$ when the P/E ratio is near to its standard value. Again, when P/E value found to be much lower than its standard value, $65\%$ of total stocks under BSE performed average and S/R value also found to be average. So 0.6 degree of belief is assigned towards the hypothesis $\{\text{\emph{Average Performance}}\}$ when P/E ratio is lower than its standard value. Similarly, when the value of P/E ratio was much higher than the standard value then it was found that $70\%$ of stocks under BSE performed poor and S/R value was also very high. So 0.7 degree of belief is assigned towards the hypothesis $\{\text{\emph{Poor Performance}}\}$. In the same way initial believes or Basic Probabilities are assigned for all other factors as mentioned in Table \ref{tab:BPA}.

\begin{table}[h]
	\small
\caption{Basic Probability Assignment}
\begin{center}
\label{tab:BPA}
\begin{tabular}{ | >{\centering\arraybackslash}m{2cm} | >{\centering\arraybackslash}m{2cm} | >{\centering\arraybackslash}m{2.15cm} | >{\centering\arraybackslash}m{2.15cm} | >{\centering\arraybackslash}m{2.15cm} | }
\hline
\bf Factors & \bf Linguistic Values & \bf High Performance (H\_P) & \bf Average Performance (A\_P) & \bf Poor Performance (P\_P)\\
\hline
\multirow {3}{1cm}{P/E}
& Low & { } & 0.6 & { }\\ \cline{2-5}
& Standard & 0.75 & { } & { }\\ \cline{2-5}
& High & { } & { } & 0.7\\ \cline{2-5}
\hline
\multirow {3}{1cm}{P/B}
& Low & { } & { } & {0.8}\\ \cline{2-5}
& Standard & { } & {0.6} & { }\\ \cline{2-5}
& High & {0.65} & { } & { }\\ \cline{2-5}
\hline
\multirow {3}{1cm}{P/S}
& Low & { } & { } & {0.7}\\ \cline{2-5}
& Standard & { } & {0.65} & { }\\ \cline{2-5}
& High & {0.75} & { } & { }\\ \cline{2-5}
\hline
\multirow {3}{1.5cm}{LTDER}
& Low & {0.6} & { } & { }\\ \cline{2-5}
& Standard & { } & {0.75} & { }\\ \cline{2-5}
& High & { } & { } & {0.65}\\ \cline{2-5}
\hline
\end{tabular}
\end{center}
\end{table} 

Now to elaborate further let us consider the following four sample rules with their initial belief:
\begin{enumerate}[Rule 1. ]
\item IF P/E is \textbf{Low} THEN Performance will be \textbf{Average} ($m_1\left( A\_P\right) =0.6$).
\item IF P/B is \textbf{Standard} THEN Performance will be \textbf{Average} ($m_2\left( A\_P\right) =0.6$).
\item IF P/S is \textbf{High} THEN Performance will be \textbf{High} ($m_4\left( H\_P\right) =0.75$).
\item IF LTDER is \textbf{High} THEN Performance will be \textbf{Poor} ($m_6\left( P\_P\right) =0.65$).
\end{enumerate}
The above four rules represent the belief towards the performance of stocks with respect to the values of four input factors as the presence of evidences. Now consider the IF part of Rule 1 and Rule 2 as the first two evidences and $m_1$ and $m_2$ as mass functions for them respectively.

Now from Rule 1, $m_1(A\_P)=0.6$ and $m_1(\Theta)=(1-0.6)=0.4$, where, $m_1(\Theta)$ represents the degree of belief in 
the rest of the hypotheses present in the hypothesis set. Similarly, from Rule 2, $m_2(A\_P)=0.6$ and 
$m2(\Theta)=(1-0.6)=0.4$. Now these two mass functions are combined using Dempster's rule of combination (Equation 9) 
to obtain a new function $m_3$ as mentioned in Table 

\begin{table}[h]
	\small
\caption{Combination of mass considering first two evidences}
\begin{center}
\label{tab:MCF1}
\begin{tabular}{ | >{\centering\arraybackslash}m{4cm} | >{\centering\arraybackslash}m{3.25cm} | >{\centering\arraybackslash}m{3.25cm} |}
\hline
Combining $m_1$ and $m_2$ & $m_2(A\_P)=0.6$ & $m_2(\Theta)=0.4$\\
\hline
$m_1(A\_P)=0.6$ & $A\_P=0.36$ & $ A\_P=0.24$\\
\hline
$m_1(\Theta)=0.4$ & $A\_P=0.24$ & $\Theta=0.16$\\
\hline
\end{tabular}
\end{center}
\end{table}      

The new mass value $m_3$ for the same hypothesis set is calculated as:
\begin{equation}
\begin{aligned}
\left. \begin{array}{ll}
\vspace{0.25cm}
m_3 (A\_P)&={0.36+0.24+0.24}=0.84\\
\vspace{0.25cm}
m_3(\Theta)&=0.16
\end{array}\right\}
\end {aligned}
\end{equation}

Now consider the IF part of the Rule 3 as the new evidence and $m_4$ be the new mass function.
From Rule 3, $m_4(H\_P)=0.75$ and $m_4(\Theta)=(1-0.75)=0.25$. Now again, $m_3$ and $m_4$ are combined to generate new mass $m_5$ as mentioned in Table \ref{tab:MCF2}. 
\begin{table}[h]
	\small
\caption{Combination of mass considering first three evidences}
\begin{center}
\label{tab:MCF2}
\begin{tabular}{ | >{\centering\arraybackslash}m{4cm} | >{\centering\arraybackslash}m{3.25cm} | >{\centering\arraybackslash}m{3.25cm} |}
\hline
Combining $m_3$ and $m_4$ & $m_4(H\_P)=0.75$ & $m_4(\Theta)=0.25$\\
\hline
$m_3(A\_P)=0.84$ & $\Phi=0.63$ & $A\_P=0.21$\\
\hline
$m_3(\Theta)=0.16$ & $H\_P=0.12$ & $\Theta=0.04$\\
\hline
\end{tabular}
\end{center}
\end{table}

New mass values $m_5$ for the hypothesis set are calculated as below:
\begin{equation}
\begin{aligned}
\left. \begin{array}{ll}
\vspace{0.25cm}
m_5 (A\_P)&=\frac{0.21}{1-0.63}=0.567\\
\vspace{0.25cm}
m_5(H\_P)&=\frac{0.12}{1-0.63}=0.324\\
m_5(\Theta)&=\frac{0.04}{1-0.63}=0.108
\end{array}\right\}
\end {aligned}
\end{equation} 

Now if we consider the IF part of the Rule 4 as the last evidence and $m_6$ as its mass function then we get $m_6(H\_P)=0.6$ and $m_6(\Theta)=(1-0.6)=0.4$. Now finally, $m_5$ and $m_6$ are combined as mentioned in Table \ref{tab:MCF3} to generate the final mass values $m_f$ for the hypothesis set (Equation 14).

\begin{table}[h]
	\small
\caption{Combination of mass considering all four evidences}
\begin{center}
\label{tab:MCF3}
\begin{tabular}{ | >{\centering\arraybackslash}m{4cm} | >{\centering\arraybackslash}m{3.25cm} | >{\centering\arraybackslash}m{3.25cm} |}
\hline
Combining $m_5$ and $m_6$ & $m_6(P\_P)=0.65$ & $m_6(\Theta)=0.35$\\
\hline
$m_5(A\_P)=0.567$ & $\Phi = 0.368$ & $A\_P=0.198$\\
\hline
$m_5(H\_P)=0.324$ & $\Phi=0.21$ & $H\_P=0.113$\\
\hline
$m_5(\Theta)=0.108$ & $P\_P=0.07$ & $\Theta=0.037$\\
\hline
\end{tabular}
\end{center}
\end{table}

\begin{equation}
\begin{aligned}
\left. \begin{array}{ll}
\vspace{0.25cm}
m_f (H\_P)&=\frac{0.113}{1-0.578}=0.267\\
\vspace{0.25cm}
m_f(A\_P)&=\frac{0.198}{1-0.578}=0.469\\
\vspace{0.25cm}
m_f(P\_P)&=\frac{0.07}{1-0.578}=0.165\\
\vspace{0.25cm}
m_f(\Theta)&=\frac{0.37}{1-0.578}=0.087
\end{array}\right\}
\end {aligned}
\end{equation} 

Following the same procedure final mass values for the rest 80 rules are calculated. For the hypothesis \emph{High Performance(H\_P)} maximum and minimum mass values ($m_f$) are found to be $0.9916$ and $0$ respectively. As mass values for any hypothesis can lie between $0$ and $1$ we have divided the favourability of any stock, based on the $m_f(H\_P)$ of the rules, into three categories: \emph{Highly Favourable} $(0.76\leq m_f(H\_P)\leq 1)$, \emph{Moderately Favourable} $(0.46\leq m_f(H\_P)\leq 0.75)$ and \emph{Not Favourable} $(m_f(H\_P)\leq 0.45)$.

So the combined version of the four sample rules mentioned earlier is formed as:

``IF P/E is \textbf {Low} AND P/B is \textbf{Standard} AND P/S is \textbf {High} AND LTDER is \textbf {High} THEN the 
stock is \textbf{Not Favourable}".

In this way total 81 fuzzy rules are formulated using DS evidence theory to develop the knowledge base of the proposed DS-fuzzy inference system and the above mentioned three selection categories are converted into fuzzy linguistic variables with trapezoidal membership values as shown in Table \ref{tab:MFOUTPUT} for output of the proposed inference system. 

\begin{table}[h]
	\small
\caption{Degree of membership of linguistic values for the output variable}
\begin{center}
\label{tab:MFOUTPUT}
\begin{tabular}{ | >{\centering\arraybackslash}m{3.5cm} | >{\centering\arraybackslash}m{3cm} | >{\centering\arraybackslash}m{4cm} |}
\hline
\bf Output Variable & \bf Linguistic Values & \bf Fuzzy Trapezoidal Membership\\
\hline
\multirow {3}{3cm}{\centering Selection (Range: 0 to 1)} & Not Favourable & (0 \hspace{0.12cm} 0 \hspace{0.12cm} 0.172 \hspace{0.12cm} 0.448) \\ \cline{2-3}
& Moderately Favourable &(0.34 \hspace{0.12cm} 0.46 \hspace{0.12cm} 0.57 \hspace{0.12cm} 0.75)\\ \cline{2-3}
& Highly Favourable & (0.64 \hspace{0.12cm} 0.88 \hspace{0.12cm} 1 \hspace{0.12cm} 1)\\ \cline{2-3}
\hline
\end{tabular}
\end{center}
\end{table} 

\begin{figure}[h]
\centering
\includegraphics[width=9cm,height=6cm]{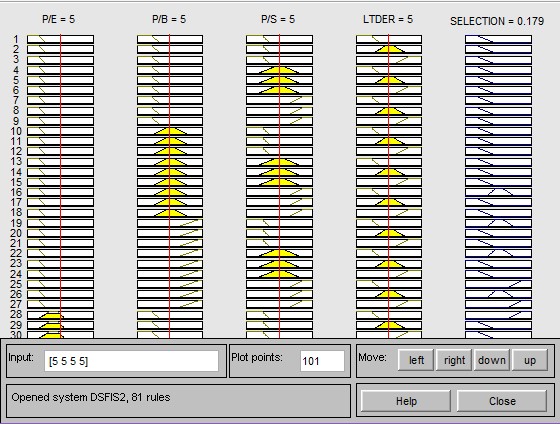}
\caption{Fuzzy rule base of the DS-fuzzy system for Stock Selection}
\label{fig:F3}
\end{figure}

\begin{figure}[h]
\centering
\includegraphics[width=9cm,height=6cm]{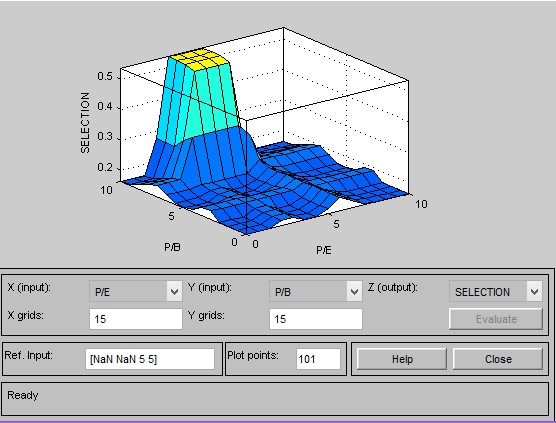}
\caption{The Sensitive analysis of the DS-fuzzy system for Stock Selection}
\label{fig:F4}
\end{figure}

Figure \ref{fig:F3} shows the snapshot of the fuzzy rule base and Figure \ref{fig:F4} shows the sensitive analysis for the same when designed in MATLAB. 

\subsubsection{Defuzzification}
The process of obtaining a single value that represent the best outcome of the fuzzy set evaluation is known as defuzzification \cite{Ngai2003}. Though there are many other popular defuzzification methods namely, Bi-sector, Mean of Maximum (MOM), Smallest of Maximum (SOM), Largest of Maximum (LOM), in this proposed model `Centroid' is used for the defuzzification purpose due to its wide range of acceptability. This method was introduced by Sugeno in 1985 and it can be expressed as:
\begin{equation}
\begin{aligned}
z^*=\frac{\int \mu_{\utilde{C}}(z)\cdot zdz}{\int \mu_{\utilde{C}}(z)dz}
\end{aligned}
\end{equation}
Where  $z$ is the output variable, $z^*$ is the defuzzified output value and $\mu_{\utilde{C}}(z)$ is the aggregated membership function.

\subsubsection{Ranking of stocks}
The proposed DS-fuzzy system is used to rank all 30 registered stocks under BSE based on the defuzzified outputs when the values corresponding to each of the four input factors for 2012-13 for each stocks are provided as input to the system. Top 10 stocks from this ranking are mentioned in Table \ref{tab:Top_Stocks} 

\begin{table}[h]
	\small
\caption{Top 10 Stocks based on the input data for FY 2012-13}
\begin{center}
\label{tab:Top_Stocks}
\begin{tabular}{ | >{\centering\arraybackslash}m{2cm} | >{\centering\arraybackslash}m{5cm} | >{\centering\arraybackslash}m{3.5cm} |}
\hline
\bf Sl. No & \bf Name of the Stocks & \bf Defuzzified Values(FY 2012-13)\\
\hline
{1} & {Hindustan Unilever Ltd.} & {0.8644}\\
\hline
{2} & {Sun Pharmaceutical Inds. Ltd.} & {0.8457}\\
\hline
{3} & {I T C Ltd.} & {0.5369}\\
\hline
{4} & {Coal India Ltd.} & {0.534}\\
\hline
{5} & {Tata Consultancy Services Ltd.} & {0.4936}\\
\hline
{6} & {Infosys Ltd.} & {0.3842}\\
\hline
{7} & {Dr. Reddy'S Laboratories Ltd.} & {0.3553}\\
\hline
{8} & {Bajaj Auto Ltd.} & {0.3292}\\
\hline
{9} & {Hero Motocorp Ltd.} & {0.3002}\\
\hline
{10} & {Cipla Ltd.} & {0.285}\\
\hline
\end{tabular}
\end{center}
\end{table}

As the consequent of every rule indicates the favourability of the stocks, higher defuzzified value indicate higher favorability. The highest defuzzified value of $0.8691$ is found for \emph{Hindustan Unilever Ltd.} among all 30 registered stocks and it has topped the ranking. These 10 stocks are used for portfolio construction as discussed in the next section.

\section{Portfolio Construction}
The first phase of stock portfolio selection, i.e. the selection of suitable stocks, is achieved in the previous section. In this section a portfolio construction model is discussed to determine optimum investment ratios for the selected securities such that the overall return is maximized under a tolerable risk. 
\subsection{Objective Function}
The ratio of the difference of fuzzy portfolio return and the risk free return to weighted mean semi-variance of the assets is used as the objective function in this work. Certainly, higher value of this ratio will indicate the better investment; so the optimization target will be to maximize the ratio.

The notations used for the construction of this objective function are mentioned below:
\begin{itemize}
\item $x_i$: Fraction of the total investment allotted to the $i^{th}$ asset;
\item $\tilde{r_i}$: Fuzzy return of the $i^{th}$ asset;
\item $r_f$: Risk free return rate;
\item $\mu_s$: Weighted mean of asset semi-variances;
\item $r_p$: Portfolio return;
\item $v_p$: Variance of the portfolio;
\item $s_p$: Skewness of the portfolio;
\end{itemize} 
Thus the objective function is formed as:
\begin{equation}
\begin{aligned}
Maximize \frac{E(\sum \tilde{r_i}x_i)-r_f}{\mu_s}
\end{aligned}
\end{equation} 
Where, after descending sort of portfolio, $\mu_s=\sum x_i s_i$, i.e. $x_i$ is the $i^{th}$ weight in the descending order and $s_i$ is the semi-variance of the $i^{th}$ ranked asset.

A fuzzy aggregation is used to calculate the fuzzy returns of the securities from the statistical database of past five years(2008-09 to 2012-13). If $t_i$ means $i^{th}$ position of the data, the fuzzy return can be calculated as:
\begin{equation}
\begin{aligned}
\tilde{r_i}=\left( { min(r_i),\frac{\sum t_i r_i}{\sum t_i},max(r_i)}\right) 
\end{aligned}
\end{equation}

The constraints included in this model are:

\begin{equation}
\begin{aligned}
\left. \begin{array}{ll}
\vspace{0.25cm}
r_p>\alpha, v_p>\beta, s_p>\gamma\\
\sum\limits_{i=1}^{n}x_i=1, x_i\leq M, x_i>0, \forall i
\end{array}\right\}
\end {aligned}
\end{equation}
Values for $\alpha$,$\beta$, $\gamma$, $M$ and $m$ are decided based on the investor's preferences. 

Thus the final portfolio optimization model can be summarized as below:
\begin{equation}
\begin{aligned}
\left. \begin{array}{ll}
\vspace{0.25cm}
Maximize \frac{E(\sum \tilde{r_i}x_i)-r_f}{\mu_s}\\
\text{Subject to, }\\
\vspace{0.25cm}
r_p>\alpha, v_p>\beta, s_p>\gamma\\
\vspace{0.5cm}
\sum\limits_{i=1}^{n}x_i=1, x_i\leq M, x_i>0, \forall i
\end{array}\right\}
\end {aligned}
\end{equation}

\subsection{Optimization of the proposed model using ACO}
An algorithm using Ant Colony Optimization (ACO) is proposed and implemented in this section to solve the optimization model as mentioned in Equation (19). ACO is a very popular meta-heuristic optimization technique inspired by the foraging behaviour of biological ants \cite{Dorigo2006,Deneubourg1990}. Pseudo code of the proposed algorithm is given as below:

\begin{algorithm}
\caption{ACO algorithm for portfolio optimization}
\label{ACOportfolio}
\begin{algorithmic}[1]
\Procedure {ACO-Portfolio}{}
\State Generate N random solution nodes based on Equation (19);
\State Initialization of the ACO;
\For {ITERATION=1 to I}
\For {ANT=1 to $\Gamma$}
\State Select the start node randomly;
\For {LIFETIME=2 to L}
\State \begin{varwidth}[t]{10cm}
Select the next node based on the heuristic information and pheromone concentration in the path. Move to the next node only if it is better than the current node;\\
\end{varwidth}
\State Update pheromone on the selected path;
\EndFor
\State Store the details of the final node reached by each ants;
\EndFor
\State 
\begin{varwidth}[t]{10.5cm}
Identify the solution node where maximum number of ants reached and consider that to be the optimum solution for the current solution;\\
\end{varwidth} 

\State 
\begin{varwidth}[t]{10.75cm}
Update the pheromone on the path of each ants who have reached this optimum solution;\\
\end{varwidth}
\State Evaporate the pheromone from all paths.
\EndFor

\EndProcedure
\end{algorithmic}
\end{algorithm}

To execute the above algorithm in this work, initially 2000 random solutions are generated for the ant colony of 50 ants. Total 400 iterations are used for the optimization purpose while lifetime for each ant is considered as 20. 

In Equation (19), $\tilde{r_i}$ is represented as a triangular fuzzy number. The Expected Return (E), Variance (Var), Skewness (Skew) and Semi-variances of these top 10 securities for past five years (2008-09 to 2012-13) as used in the above algorithm are calculated by the following theorem and are mentioned in Table \ref{tab:ER_V_Sk_S} 
\newtheorem{theorem}{Theorem}[section]
\begin{theorem}
Let $\tilde{A}=(a,b,c)$ be a triangular fuzzy number. The  the weighted possibilistic mean, variance and skewness can be calculated as \cite{Bhattacharyya2013}:
\begin{equation}
\begin{aligned}
\left. \begin{array}{ll}
\vspace{0.25cm}
E(\tilde{A})&=\frac{1}{6}(a+4b+c)\\
\vspace{0.25cm}
Var(\tilde{A})&=\frac{1}{18}(a^2+b^2+c^2-ab-bc-ca)\\
\vspace{0.25cm}
Skew(\tilde{A})&=\frac{19(a^3+c^3)-8b^3-42b(a^2+c^2)+12b^2(a+c)-15(a^2c+ac^2)+60abc}{10\sqrt{2}(\sqrt{a^2+b^2+c^2-ab-bc-ca})^3}
\end{array}\right\}
\end {aligned}
\end{equation}
\end{theorem}  

\begin{table}[h]
	\small
\caption{Expected Return, Variance, Skewness and Semivariance of stocks}
\begin{center}
\label{tab:ER_V_Sk_S}
\begin{tabular}{ | >{\centering\arraybackslash}m{0.75cm} | >{\centering\arraybackslash}m{3cm} | 
>{\centering\arraybackslash}m{1.5cm}|>{\centering\arraybackslash}m{1.5cm}|>{\centering\arraybackslash}m{1.5cm}|>{\centering\arraybackslash}m{2cm}|}
\hline
Rank &  Name of the Stocks &  Return &  Variance &  Skewness &  Semivariance\\
\hline
{1}& {Hindustan Unilever Ltd.} & {0.1443} & {0.0001} & {0.2391}  & {0.00004}\\
\hline
{2} & {Sun Pharmaceutical Inds. Ltd.}& {0.0529} & {0.0006} & {-0.3269} & {0.00254}\\
\hline
{3} & {I T C Ltd.} & {0.2801} & {0.0035} & {0.7523} & {0.00314}\\
\hline
{4} & {Coal India Ltd.} & {0.1275} & {0.0028} & {0.6046} & {0.0373}\\	
\hline
{5} & {Tata Consultancy Services Ltd.} & {0.1228} & {0.0013} & {-0.5029} & {0.00557}\\		
\hline
{6} & {Infosys Ltd.} & {0.0356} & {0.0003} & {0.8648} & {0.0002}\\		
\hline
{7} & {Dr. Reddy'S Laboratories Ltd.} & {0.0484} & {0.0006} & {-0.0065} & {0.00098}\\		
\hline
{8} & {Bajaj Auto Ltd.} & {0.0579} & {0.001} & {0.7262} & {0.0013}\\		
\hline
{9} & {Hero Motocorp Ltd.} & {0.1379} & {0.0004} & {0.8057} & {0.00049}\\		
\hline
{10} & {Cipla Ltd.} & {0.0437} & {0.0002} & {0.8258} & {0.00018}\\		
\hline
\end{tabular}
\end{center}
\end{table}

When the algorithm is executed in MATLAB with the above dataset and considering other parameters as $r_f=0.01$, $ \beta 
=0.5 $, $\alpha =0.05$, $ \gamma =0.001$, $ M=0.8 $ and $ \mu_s=0.0016 $, the maximum return is found as $0.1317$. The 
proposed ratio allocation for this return is given in Table \ref{tab:Ratio_allocation} .

\begin{table}[h]
	\small
	\caption{Ratio allocation for the proposed portfolio}
	\begin{center}
		\label{tab:Ratio_allocation}
		\begin{tabular}{| >{\centering\arraybackslash}m{1.5cm} | >{\centering\arraybackslash}m{1cm} | 
				>{\centering\arraybackslash}m{0.75cm}| >{\centering\arraybackslash}m{1cm}| 
				>{\centering\arraybackslash}m{0.75cm}| >{\centering\arraybackslash}m{1cm}| 
				>{\centering\arraybackslash}m{1cm}| >{\centering\arraybackslash}m{1cm}| 
				>{\centering\arraybackslash}m{1cm}| >{\centering\arraybackslash}m{0.75cm}|}
			\hline 
			Hindustan Unilever Ltd. & Sun Pharma. Inds. Ltd. & ITC Ltd. & Coal India Ltd. & TCS Ltd. & Infosys Ltd. & 
			Dr. Reddy's Lab. Ltd. & Bajaj Auto Ltd. & Hero Motocorp Ltd. & Cipla Ltd. \\
			\hline
			0.2403 & 0.2235 & 0.1970 & 0.0764 & 0.0574 & 0.0525 & 0.0452 & 0.0413 & 0.0407 & 0.0255 \\
		\hline
		\end{tabular}
		\end{center}
\end{table}

\begin{figure}[h]
	\centering
	\includegraphics[width=9cm,height=6cm]{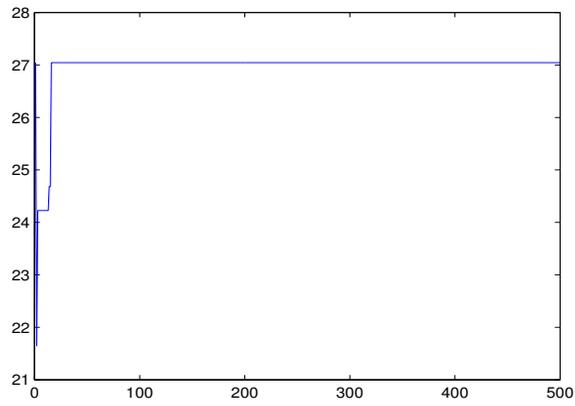}
	\caption{Convergence of objective values based on proposed ranking}
	\label{fig:F5}
\end{figure}

\begin{figure}[h]
	\centering
	\includegraphics[width=9cm,height=6cm]{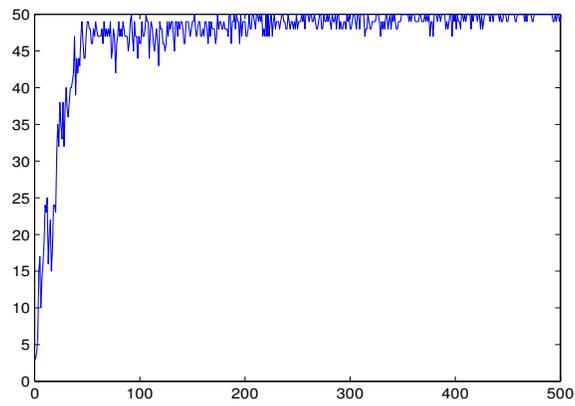}
	\caption{Ant accumulation at optimum solutions}
	\label{fig:F6}
\end{figure}

The convergence of the objective values as per the proposed model is depicted in Figure \ref{fig:F5} and the 
accumulation of ants to the optimum objective values in each iteration is depicted in Figure \ref{fig:F6}. 

\section{Result Analysis}
\noindent To justify the applicability of the proposed model its performance is analyzed further in three aspects. 
First, top 15 companies are ranked based of their S/R values for the year of 2013-14 and 2014-15. Details of these 
rankings along with the ranking of the proposed model is given in Table \ref{tab:rank_comparison}.

\begin{table}[htbp]
	\centering
	\caption{Top 15 Stocks}
	\begin{tabular}{| >{\centering\arraybackslash}m{1cm} | >{\centering\arraybackslash}m{3cm} | 
							>{\centering\arraybackslash}m{3cm}| >{\centering\arraybackslash}m{3cm}|}
		\hline 
		Rank  & Stocks Ranking based on the proposed model  & Stocks Ranking based on the S/R values (2013-14) & Stocks 
		Ranking based on the S/R values (2014-15) \\
		\hline 
		1     & Hindustan Unilever Ltd. & Sesa Sterlite Ltd. & State Bank Of India \\ \hline 
		2     & Sun Pharmaceutical Inds. Ltd & NTPC Ltd. & Infosys Ltd. \\ \hline 
		3     & ITC Ltd. & Hero Motocorp Ltd. & ITC Ltd. \\ \hline 
		4     & Coal India Ltd. & Maruti Suzuki India Ltd. & HDFC Bank Ltd. \\ \hline 
		5     & TCS Ltd. & Hindustan Unilever Ltd. & Hero Motocorp Ltd. \\ \hline 
		6     & Infosys Ltd. & Cipla Ltd. & Hindustan Unilever Ltd. \\ \hline 
		7     & Dr. Reddy's Laboratories Ltd. & State Bank Of India & Wipro Ltd. \\ \hline 
		8     & Bajaj Auto Ltd. & Bharat Heavy Electricals Ltd. & Maruti Suzuki India Ltd. \\ \hline 
		9     & Hero Motocorp Ltd. & Wipro Ltd. & Tata Power Co. Ltd. \\ \hline 
		10    & Cipla Ltd. & ITC Ltd. & Tata Motors Ltd. \\ \hline 
		11    & HDFC Bank Ltd. & Tata Power Co. Ltd. & TCS Ltd. \\ \hline 
		12    & Wipro Ltd. & Infosys Ltd. & Coal India Ltd. \\ \hline 
		13    & Larsen \& Toubro Ltd. & Hindalco Industries Ltd. & Mahindra \& Mahindra Ltd. \\ \hline 
		14    & Bharti Airtel Ltd. & Dr. Reddy'S Laboratories Ltd. & Cipla Ltd. \\ \hline 
		15    & Tata Power Co. Ltd. & Sun Pharmaceutical Inds. Ltd. & Sun Pharmaceutical Inds. Ltd. \\ \hline 
			\end{tabular}%
	\label{tab:rank_comparison}%
\end{table}%

\noindent From the above table we can find a match of 9 companies for FY 2013-14 and a match of 11 companies for FY 
2014-15. As S/R values are effective performance indicator of stocks, it can be concluded that the proposed model will 
give better return in short-term investment period. 

In the second stage by considering top $10$ stocks under BSE, based on their S/R values for FY 2012-13, a 
portfolio is constructed using the same ACO algorithm and the objective function (Eq. 19). The optimum return is found 
to be $0.0740$, which is much less than the expected return ($0.1317$) of the proposed model. 

\begin{figure}[h]
	\centering
	\includegraphics[width=9cm,height=6cm]{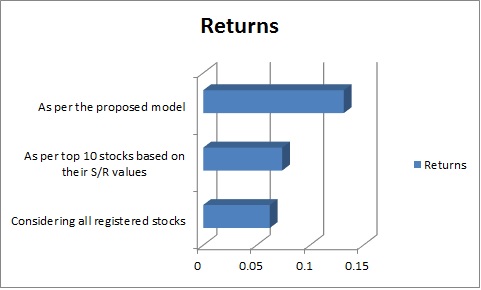}
	\caption{Portfolio returns}
	\label{fig:F7}
\end{figure}

In the third stage another portfolio is constructed using the same ACO algorithm, where all 30 registered companies 
were under BSE are considered. But the optimum return found to be $0.0624$ which is much lower than the return 
calculated by the proposed model. This again proves that the proposed model is more effective in portfolio construction.
Figure \ref{fig:F7} compares the returns of these three portfolios.

\section{Conclusion}
\noindent In this research a novel fuzzy expert system model is designed and implemented to evaluate and rank the 
stocks under BSE. DS-evidence theory is used for the development of the fuzzy rule base to reduce the overall 
implementation time and cost of the system. A portfolio optimization model is constructed by considering the ratio of 
the difference of fuzzy portfolio return and risk free return to the weighted mean semi-variance of the assets as the 
objective function. ACO is used to obtain the optimum value of this portfolio. 

As the outcome of this model found to be satisfactory, this can be implemented for any stock exchanges around the world 
but the selection of critical factors may vary over different stock exchanges. This fuzzy expert system model can be 
used to rank any set of alternatives based on the factors influencing them. The portfolio optimization model and the  
related ACO algorithm can be used to optimize any rank-preference based portfolios.     
 
\bibliographystyle{elsarticle-num}
\bibliography{BIBPORTFOLIO}

\end{document}